%% file: RobICML.tex
\documentclass{article}

\pdfoutput=1

\usepackage{times,amsmath,amsfonts,amssymb,url,color,graphicx}

\usepackage{subfigure} 

\usepackage{natbib}

\usepackage{boxedminipage}

\usepackage[accepted]{icml2016}

\usepackage{tikz}
\usepackage{pgfplots}

\newcommand{\simp}{{\mathcal{S}}}
\newcommand{\Loss}{{\Lambda}}

\newcommand{\X}{{\mathcal{X}}}
\newcommand{\Y}{{\mathcal{Y}}}
\newcommand{\D}{{\mathcal{D}}}
\newcommand{\cW}{{\mathcal{W}}}

\newcommand{\cH}{{\mathcal H}}

\newcommand{\sign}{{\mathrm {sign}}}

\newcommand{\avg}{{\mathrm{avg}}}

\newtheorem{lemma}{Lemma}

\newtheorem{theorem}{Theorem}
\newtheorem{corollary}{Corollary}
\newtheorem{definition}{Definition}
\newtheorem{example}{Example}

\newcommand{\BlackBox}{\rule{1.5ex}{1.5ex}}  % end of proof
\newenvironment{proof}{\par\noindent{\bf Proof\ }}{\hfill\BlackBox\\[2mm]}

\DeclareMathOperator*{\E}{\mathbb{E}}
\DeclareMathOperator*{\prob}{\mathbb{P}}

\newcommand{\reals}{\mathbb{R}}

\newcommand{\figref}[1]{Figure~\ref{#1}}

\newcommand{\secref}[1]{Section~\ref{#1}}
\newcommand{\appref}[1]{Appendix~\ref{#1}}

\newcommand{\thmref}[1]{Theorem~\ref{#1}}

\newcommand{\lemref}[1]{Lemma~\ref{#1}}

\newcommand{\exampleref}[1]{Example~\ref{#1}}

\newcommand{\inner}[1]{\langle #1\rangle}

\newcommand{\figurespath}[1]{figures/}

\newenvironment{myalgo}[1]%
{
\begin{center}
\begin{boxedminipage}{0.8\linewidth}
\begin{center}
\textbf{\texttt{#1}}
\end{center}
\rm
\begin{tabbing}
....\=...\=...\=...\=...\=  \+ \kill
} %
{\end{tabbing} 
\end{boxedminipage} \end{center} %\vspace{0.3cm}
}

\icmltitlerunning{Minimizing the Maximal Loss: How and Why}

\begin{document}

\twocolumn[
\icmltitle{Minimizing the Maximal Loss: How and Why}

\icmlauthor{Shai Shalev-Shwartz}{shais@cs.huji.ac.il}
\icmladdress{School of Computer Science and
    Engineering, The Hebrew University, Jerusalem, Israel.}
\icmlauthor{Yonatan Wexler}{yonatan.wexler@orcam.com}
\icmladdress{Orcam}

%\icmlkeywords{boring formatting information, machine learning, ICML}

\vskip 0.3in
]

\input{Rob.tex}

\bibliography{bib}
\bibliographystyle{icml2016}

\appendix
\onecolumn

\input{RobProofs.tex}

\end{document}

%% file: Rob.tex
\begin{abstract}
  A commonly used learning rule is to approximately minimize the
  \emph{average} loss over the training set. Other learning
  algorithms, such as AdaBoost and hard-SVM, aim at minimizing the
  \emph{maximal} loss over the training set. The average loss is more
  popular, particularly in deep learning, due to three main
  reasons. First, it can be conveniently minimized using online
  algorithms, that process few examples at each iteration. Second, it
  is often argued that there is no sense to minimize the loss on the
  training set too much, as it will not be reflected in the
  generalization loss.  Last, the maximal loss is not robust to
  outliers. In this paper we describe and analyze an algorithm that
  can convert any online algorithm to a minimizer of the maximal
  loss. We prove that in some
  situations better accuracy on the training set is crucial to obtain
  good performance on unseen examples. Last, we propose robust
  versions of the approach that can handle outliers.
\end{abstract}

\section{Introduction}

In a typical supervised learning scenario, we have training examples,
$S = ((x_1,y_1),\ldots,(x_m,y_m)) \in (\X \times \Y)^m$, and our goal
is to learn a function $h : \X \to \Y$. We focus on the case in which
$h$ is parameterized by a vector $w \in \cW \subset \reals^d$, and
we use $h_w$ to denote the function induced by $w$. The performance
of $w$ on an example $(x,y)$ is assessed using a loss function, $\ell
: \cW \times \X \times \Y \to [0,1]$. A commonly used learning rule
is to approximately minimize the average loss, namely,
\begin{equation} \label{eqn:avgGoal}
\min_{w \in \cW} L_\avg(w) := \frac{1}{m}\sum_{i=1}^m \ell(w,x_i,y_i) ~.
\end{equation}
Another option is to approximately minimize the maximal loss, namely,
\begin{equation} \label{eqn:maxGoal}
\min_{w \in \cW} ~ L_{\max}(w) := \max_{i \in [m]} \ell(w,x_i,y_i) ~.
\end{equation}
Obviously, if there exists $w^* \in \cW$ such that
$\ell(w^*,x_i,y_i)=0$ for every $i$ then the minimizers of both
problems coincide. However, approximate solutions can be very
different. In particular, since $L_{\max}(w) \ge L_\avg(w)$ for every
$w$, the guarantee
$L_{\max}(w) < \epsilon$ is stronger than the guarantee $L_\avg(w)
< \epsilon$. Furthermore, for binary classification with the
zero-one loss, any vector for which $L_{\max}(w) < 1$ must predict all
the labels on the training set correctly, while the guarantee
$L_\avg(w) < 1$ is meaningless. 

Some classical machine learning algorithms can be viewed as
approximately minimizing $L_{\max}$. For example, 
% many boosting
% algorithms can be viewed as coordinate descent algorithms for solving
% $L_{\max}$ with respect to the loss $\ell(w,x_i,y_i) = -y_i
% \inner{w,x_i}$ (here $\inner{w,x_i}$ is the inner product of $w$
% and $x_i$) and over the set $\cW$ being the probabilistic simplex
% (see for example \cite{shalev2010equivalence} and the references
% therein). 
Hard-SVM effectively solves $L_{\max}$ with
respect to the loss function $\ell(w,x_i,y_i) = \lambda \|w\|^2 +
1[y_i \inner{w,x_i} < 1]$. 
However, minimizing $L_\avg$ is a more popular approach, especially
for deep learning problems, in which $w$ is the vector of weights of a
neural network and the optimization is performed using variants of
stochastic gradient descent (SGD). There are several reasons to prefer
$L_\avg$ over $L_{\max}$:
\begin{enumerate}
\item If $m$ is very large, it is not practical to perform operations
  on the entire training set. Instead, we prefer iterative algorithms that
  update $w$ based on few examples at each iteration. This can be
  easily done for $L_\avg$ by observing that if we sample $i$
  uniformly at random from $[m]$, then the gradient of
  $\ell(w,x_i,y_i)$ with respect to $w$ is an unbiased estimator of
  the gradient of $L_\avg(w)$. This property, which lies at the heart
  of the SGD algorithm, does not hold for $L_{\max}$. 
\item Our ultimate goal is not to minimize the loss on the training
  set but instead to have a small loss on unseen examples.  As argued
  before, approximately minimizing $L_{\max}$ can lead to a smaller loss on the
  training set, but it is not clear if this added accuracy will also
  be reflected in performance on unseen examples. Formal arguments of
  this nature were given in
  \cite{bousquet2008tradeoffs,shalev2008svm}.
% In particular, in
%   the usual statistical learning model (see \cite{MLbook} for an
%   overview), we assume that the examples in the training set are
%   sampled i.i.d. from an underlying distribution $\D$ over $\X \times
%   \Y$, and our ultimate goal is to minimize $L(w):=\E_{(x,y) \sim
%     \D}[\ell(w,x,y)]$. The objective $L_\avg(w)$ is a proxy for
%   $L(w)$, but due to the finiteness of $m$ there will always be a gap
%   between the two.
\item The objective $L_{\max}$ is not robust to outliers. It is easy
  to see that even a single outlier can make the minimizer of
  $L_{\max}$ meaningless. 
\end{enumerate}

In this paper we tackle the aforementioned disadvantages of
$L_{\max}$, and by doing so, we show cases in which $L_{\max}$ is
preferable. In particular:
\begin{enumerate}
\item We describe and analyze a meta algorithm that can take any
  online learner for $w$ and convert it to a minimizer of
  $L_{\max}$. 
% In particular, we can take online gradient descent as
%   the online learner, and obtain a variant of SGD for minimizing
%   $L_{\max}$. We prove that the number of iterations required by our
%   meta algorithm is upper bounded by $O(m \log m + B(\epsilon))$,
%   where $B(\epsilon)$ is the number of iterations required by the
%   online algorithm to achieve error rate of $\epsilon$. In particular,
%   for the online gradient descent variant, the bound should be
%   compared to the bound $B(\epsilon)$ of SGD, which usually has the
%   form $B(\epsilon) = C/\epsilon^p$ for some $p \in [1,2]$ and for
%   some $C \in \reals$. This means that we pay an additive term of $m
%   \log(m)$. To understand our gain, recall that the main reason to
%   prefer $L_{\max}$ over $L_\avg$ is that for the same value of
%   $\epsilon$, the guarantee $L_{\max}(w) < \epsilon$ is much stronger
%   than $L_\avg(w) < \epsilon$. In the example of binary classification
%   with the hinge-loss, to obtain a perfect predictor we need to set
%   $\epsilon < 1$ for $L_{\max}$ but we need to set $\epsilon < 1/m$
%   for $L_\avg$. This means that the overall runtime for this example
%   is $O(m \log(m) + C)$ for $L_{\max}$ and $O(m^p C)$ for
%   $L_\avg$. Since $C$ can be quite large (it measures the capacity of
%   the hypothesis class), this different is very significant even if
%   $p=1$. 
A detailed description of our meta algorithm, its analysis, and a
comparison to other approaches, are given in \secref{sec:algo}. 
\item The arguments in \cite{bousquet2008tradeoffs,shalev2008svm} rely
  on a comparison of upper bounds. We show that these upper bounds are
  not tight in many cases. Furthermore, we analyze the sample
  complexity of learning in situations where the training examples are
  divided to ``typical'' scenarios and ``rare'' scenarios. We argue
  that in many practical cases, our goal is to have a high accuracy on
  both typical and rare examples. We show conditions under which
  minimizing even few rare examples suffice to guarantee good
  performance on unseen examples from the rare scenario. In other
  words, few examples can have a dramatic effect on the performance of the learnt classifier
  on unseen examples. This is described and
  analyzed in \secref{sec:why}. 
\item Finally, in \secref{sec:robust} we review standard techniques
  for generalizing the results from realizable cases to scenarios in
  which there might be outliers in the data.
\end{enumerate}

To summarize, we argue that in some situations minimizing $L_{\max}$
is better than minimizing $L_\avg$. We address the ``how'' question in
\secref{sec:algo}, the ``why'' question in \secref{sec:why}, and the
issue of robustness in \secref{sec:robust}. 
Finally, in
\secref{sec:experiments} we provide some empirical evidence, showing
the effectiveness of our algorithm on real world learning problems.

\section{How} \label{sec:algo}

In this section we describe and analyze an algorithmic framework for
approximately solving the optimization problem given in
\eqref{eqn:maxGoal}. 

Denote by $\simp_m = \{ p \in [0,1]^m : \|p\|_1 = 1\}$ the
probabilistic simplex over $m$ items. We also denote by $\Loss : \cW
\to [0,1]^m$ the function defined by
\[
\Loss(w) = (\ell(w,x_1,y_1),\ldots,\ell(w,x_m,y_m)) ~.
\] 
The first step is to note that the optimization problem given in \eqref{eqn:maxGoal} is
equivalent to 
\begin{equation} \label{eqn:pLoss}
\min_{w \in \cW} ~ \max_{p \in \simp_m} ~ \inner{p, \Loss(w)} ~.
\end{equation}
This is true because for every $w$, the $p$ that maximizes the inner
optimization is the all zeros vector except $1$ in the coordinate for
which $\ell(w,x_i,y_i)$ is maximal. 

We can now think of \eqref{eqn:pLoss} as a zero-sum game between
two-players. The $p$ player tries to maximize $\inner{p,\Lambda(w)}$ while
the $w$ player tries to minimize $\inner{p,\Lambda(w)}$. 
The optimization process is comprised of $T$ game rounds. At round $t$,
the $p$ player defines $p_t \in \simp_m$ and the $w$ player defines
$w_t \in \cW$. We then sample $i_t \sim p_t$ and define the value of the round to be
$\ell(w_t,x_{i_t},y_{i_t})$. 

To derive a concrete algorithm we need to specify how player $p$ picks
$p_t$ and how player $w$ picks $w_t$. For the $w$ player one can use any online
learning algorithm. We specify the requirement from the algorithm
below. 

\begin{definition}[Mistake bound for the $w$ player]
We say that the $w$ player enjoys a mistake bound of
$C$ if  for every sequence of indices $(i_1,\ldots,i_T) \in [m]^T$ we have that 
\begin{equation} \label{eqn:lowReg_w}
\sum_{t=1}^T \ell(w_t,x_{i_t},y_{i_t}) ~\le~ C ~.
\end{equation}
\end{definition}

\begin{example} \label{example:1}
  Consider a binary classification problem in which the data is
  linearly separable by a vector $w^*$ with a margin of $1$. Let the
  loss function be the zero-one loss, namely, $\ell(w,x,y) = 1[y
  \inner{w,x} \le 0]$, where $1[\textrm{boolean expression}]$ is $1$
  if the boolean expression holds and $0$ otherwise. We can use the
  online Perceptron algorithm as our $w$ learner and it is well known
  that the Perceptron enjoys the mistake bound of $C =
  \|w^*\|^2\,\max_{i \in [m]} \|x_i\|^2$ (for a reference, see for
  example \cite{shalev2011online}).
\end{example}

For the $p$ player, we use the seminal work of
\cite{auer2002nonstochastic}. In particular, recall that the goal of
the $p$ player is to maximize the loss, $\ell(w_t,x_{i_t},y_{i_t}$,
where $i_t \sim p_t$. The basic idea of the construction is therefore
to think of the $m$ examples as $m$ slot machines, where at round $t$
the gain of pulling the arms of the different machines is according to
$\Lambda(w_t) \in [0,1]^m$. Crucially, the work of
\cite{auer2002nonstochastic} does not assume that $\Lambda(w_t)$ are
sampled from a fixed distribution, but rather the vectors
$\Lambda(w_t)$ can be chosen by an adversary. As observed in
\citet[Section 9]{auer2002nonstochastic}, this naturally fits zero-sum
games, as we consider here.

In \cite{auer2002nonstochastic} it is proposed to rely on the
algorithm EXP3.P.1 as the strategy for the $p$-player. The acronym
EXP3 stands for
\textbf{Exp}loration-\textbf{Exp}loitation-\textbf{Exp}onent, because
the algorithm balances between exploration and exploitation and rely
on an exponentiated gradient framework.  The ``P'' in EXP3.P.1 stands
for a regret bound that holds with high probability. This is essential
for our analysis because we will later apply a union bound over the
$m$ examples. While the EXP3.P.1 algorithm gives the desired regret
analysis, the runtime per iteration of this algorithm scales with
$m$. Here, we propose another variant of EXP3 for which the runtime
per iteration is $O(\log(m))$. 

To describe our strategy for the $p$ player, recall that it maintains
$p_t \in \simp_m$. We will instead maintain another vector, $q_t \in
\simp_m$, and will set $p_t$ to be the vector such that $p_{t,i} =
\frac{1}{2} q_{t,i} + \frac{1}{2m}$. That is, $p_t$ is a half-half mix
of $q_t$ with the uniform distribution. While in general such a strong
mix with the uniform distribution can hurt the regret, in our case it
only affects the convergence rate by a constant factor. On the up
side, this strong exploration helps us having an update step that
takes $O(\log(m))$ per iteration.

Recall that at round $t$ of the algorithm, we sample $i_t \sim p_t$
and the value of the round is $\ell(w_t,x_{i_t},y_{i_t})$. Denote $z_t
= - \frac{\ell_{i_t}(w_t)}{p_{i_t}} e_{i_t}$, then it is easy to
verify that $\E_{i_t \sim p_t}[z_t] = - \Lambda(w_t)$. Therefore,
applying gradient descent with respect to the linear function
$\inner{\cdot,z_t}$ is in expectation equivalent to applying gradient
descent with respect to the linear function
$-\inner{\cdot,\Lambda(w_t)}$, which is the function the $p$ player
aims at minimizing. Instead of gradient descent, we use the
exponentiated gradient descent approach which applies gradient descent
in the log space, namely, the update can be written as $\log(q_{t+1})
= \log(q_t) + \eta z_t$. 

A pseudo-code of the resulting algorithm is given in
\secref{sec:pseudo}. Observe that we use a tree structure to hold 
the vector $q$, and since all but the $i_t$ coordinate of $z_t$ are
zeros, we can implement the update of $q$ in $O(\log(m))$ time per
iteration. The following theorem summarizes the convergence
of the resulting algorithm. 

\begin{theorem} \label{thm:mainAlgo} Suppose the we have an oracle
  access to an online algorithm that enjoys a mistake bound of $C$
  with respect to the training examples $(x_1,y_1),\ldots,(x_m,y_m)$.
  Fix $\epsilon,\delta$, and suppose we run the FOL algorithm with
  $T,k$ such that $C/T \le \epsilon/8$, $T =
  \Omega(m\log(m/\delta)/\epsilon)$, and $k =
  \Omega(\log(m/\delta)/\epsilon)$, and with $\eta = 1/(2m)$. Then,
  with probability of at least $1-\delta$,
\[
\max_i \frac{1}{k} \sum_{j=1}^k \ell(w_{t_j},x_i,y_i)  ~\le~
\epsilon ~.
\]
\end{theorem}
The proof of the theorem is given in \appref{sec:proofMain}.

The above theorem tells us that we can find an ensemble of
$O(\log(m)/\epsilon)$ predictors, such that the ensemble loss is
smaller than $\epsilon$ for all of the examples. 

We next need to show that we can construct a single predictor with a small
loss. To do so, we consider two typical scenarios. The first is
classification settings, in which $\ell(w,x,y)$ is the zero-one
loss and the second is convex losses in which $\ell(w,x,y)$
has the form $\phi_y(h_w(x))$, where for every $y$, $\phi_y$ is a
convex function.

\subsection{Classification}
In classification, $\ell(w,x,y)$ is the zero-one loss, namely,
it equals to zero if $h_w(x)=y$ and it equals to $1$ if $h_w(x) \neq
y$. We will take $\epsilon$
to be any number strictly smaller than $1/2$, say $0.499$.

Observe that \thmref{thm:mainAlgo} tells us that the average loss of
the classifiers $w_{t_1},\ldots,w_{t_k}$ is smaller than
$\epsilon=0.499$. Since the values of the loss are either $1$ or $0$,
it means that the loss of more than $1/2$ of the classifiers is $0$, which
implies that the majority classifier has a zero loss. 
\begin{corollary}
  Assume that $\ell(w,x,y)$ is the zero-one loss function, namely,
  $\ell(w,x,y) = 1[h_w(x) \neq y]$.  Apply \thmref{thm:mainAlgo} with
  $\epsilon = 0.49$. Then, with probability of at least $1-\delta$, the
  majority classifier of $h_{w_{t_1}},\ldots,h_{w_{t_k}}$ is
  consistent, namely, it makes no mistakes on the
  entire training set. 
\end{corollary}

\begin{example}
  Consider again the linear binary classification problem given in
  \exampleref{example:1}, where we use the online
  Perceptron algorithm as our $w$ learner, and its mistake bound is
  $C$ as given in \exampleref{example:1}. Then, after
  $\tilde{O}\left(m + C \right)$ iterations, we will find an ensemble of $O(\log(m))$ halfspaces,
 whose majority vote is consistent with all the examples. In
 \secref{sec:related} we compare the runtime of the method to
 state-of-the-art approaches. Here we just note that to obtain a
 consistent hypothesis using SGD one needs order of
 $m \, C$ iterations, which is significantly larger in most scenarios. 
\end{example}

\subsection{Convex Losses}
Consider now the case in which $\ell(w,x,y)$ has the form
$\phi_y(h_w(x))$, where for every $y$, $\phi_y$ is a convex function.
Note that this assumption alone does not imply that $\ell$ is a convex
function of $w$ (this will be true only if $h_w(x)$ is an affine
function). 

In the case of convex $\phi_y$, combining \thmref{thm:mainAlgo} with
Jensen's inequality we obtain:
\begin{corollary}
Under the assumptions of \thmref{thm:mainAlgo}, if $\ell(w,x,y)$ has the form
$\phi_y(h_w(x))$, where for every $y$, $\phi_y$ is a convex function,
then the predictor
$h(x) = \frac{1}{k} \sum_{j=1}^k h_{w_{t_j}}(x)$
satisfies 
$
\forall i,~~~ \phi_{y_i}(h(x_i)) ~\le~ \epsilon ~.
$
If we further assume that $h_w(x)$ is an affine function of $w$, and
let $w =  \frac{1}{k} \sum_{j=1}^k w_{t_j}$,
then we also have that
\[
\forall i,~~~ \phi_{y_i}(h_{w}(x_i)) ~\le~ \epsilon ~.
\]
\end{corollary}

% \begin{example}
%   Consider a linear regression problem where $\X$ is the $\ell_2$ ball
%   of radius $X$, $\cW$ is the $\ell_2$ ball of radius $W$, $\Y = [0,1], 
%   h_w(x) = \max\{0,\min\{1,\inner{w,x}\}\}$, and $\ell(w,x,y) =
%   |h_w(x)-y|$. For the $w$ player we can apply the online gradient
%   descent algorithm (see \cite{shalev2011online}) with respect to the
%   convex surrogate loss $\ell^s(w,x,y) = |\inner{w,x}-y|$. The regret
%   bound of online gradient descent (see again
%   \cite{shalev2011online}) guarantees that for every sequence of
%   examples we have
% \[
% \sum_{t=1}^T \ell^s(w_t,x_{i_t},y_{i_t}) \le \sum_{t=1}^T
% \ell^s(w^*,x_{i_t},y_{i_t}) + X\, W\, \sqrt{T} ~.
% \]
%   is $R_w(T) \le X W / \sqrt{T}$. It follows that to find a
%   predictor with error of at most $\epsilon$ on all the examples we need that $T =
%   \Omega\left( \frac{ X^2 W^2}{\epsilon^2} + \frac{m \log(m)}{\epsilon}\right)$. In
%   contrast, to achieve the same result with SGD we need order of $\frac{X^2 W^2
%     m^2}{\epsilon^2}$ iterations. 
% \end{example}

\subsection{Pseudo-code} \label{sec:pseudo}

Below we describe a pseudo-code of the algorithm. We rely on a tree
data structure for maintaining the probability of the $p$-player. It
is easy to verify the correctness of the implementation. Observe that
the runtime of each iteration is the time required to perform one step
of the online learner plus $O(\log(m))$ for sampling from $p_t$ and
updating the tree structure. 

\begin{myalgo}{Focused Online Learning (FOL)}
\textbf{Input:} \+ \\
Training examples $(x_1,y_1),\ldots,(x_m,y_m) $ \\
Loss function $\ell : \cW \times \X \times \Y \to [0,1]$ \\
Parameters $\eta,T,k$ \\
Oracle access to online learning algorithm $\mathrm{OLA}$ \- \\
\textbf{Initialization:} \+ \\
$\textrm{Tree.initialize}(m)$  (see the Tree pseudo-code)\\
$w_1 = \textrm{OLA.initialize}()$ \- \\
\textbf{Loop over $t \in \{1,\ldots,T \}$:} \+ \\
$(i_t,p_{i_t}) = \textrm{Tree.sample}(1/2)$ \\
$\textrm{OLA.step}(x_{i_t},y_{i_t})$ \\
$\textrm{Tree.update}(i_t,\exp(\eta \,
\ell(w_t,x_{i_t},y_{i_t}) / p_{i_t}))$ \- \\
\textbf{Output:} \+ \\
Sample $(t_1,\ldots,t_k)$ indices uniformly from $[T]$ \\
Output Majority/Average of $(h_{w_{t_1}},\ldots,h_{w_{t_k}})$
\end{myalgo}

\begin{myalgo}{Tree}
\textbf{initialize(m)} \+ \\
Build a full binary tree of height $h={\lceil \log_2(m) \rceil}$ \\
Set value of the first $m$ leaves to $1$ and the rest to $0$ \\
Set the value of each internal node to be \+ \\ the sum of its two children
\- \\
Let $q_i$ be the value of the $i$'th leaf divided by \+ \\ the value of the
root \- \- \\
\textbf{sample($\gamma$)} \+ \\
Sample $b \in \{0,1\}$ s.t. $\prob[b=0]=\gamma$ \\
If $b=0$ \+ \\
 Sample $i$ uniformly at random from $[m]$ \- \\
Else \+ \\
 Set $v$ to be the root node of the tree \\
 While $v$ is not a leaf: \+ \\
  Go to the left/right child by sampling \+ \\ according to their
  values \- \- \\
 Let $i$ be the obtained leaf \- \\
 Return: $(i,\gamma/m+(1-\gamma)q_i)$ \- \\
\textbf{update($i,f$)} \+ \\
 Let $v$ be the current value of the $i$'th leaf of the tree \\
 Let $\delta = f \, v - v$ \\
 Add $\delta$ to the values of all nodes on \+ \\ the path from the $i$'th
 leaf to the root \- \-
\end{myalgo}

\subsection{Related Work} \label{sec:related}

As mentioned before, our algorithm is a variant of the approach given
in \citet[Section 9]{auer2002nonstochastic}, but has the advantage
that the update of the $p$ player at each iteration scales with
$\log(m)$ rather than with $m$.  Phrasing the max-loss minimization as
a two players game has also been proposed by
\cite{clarkson2012sublinear,hazan2011beating}. These works focus on
the specific case of binary classification with a linear predictor,
namely, they tackle the problem $\min_{w \in \reals^d: \|w\|_2 \le 1}
\max_{p \in \simp_m} \sum_i p_i \inner{w,x_i}$. Assuming the setup of
\exampleref{example:1}, \cite{clarkson2012sublinear} presents an
algorithm that finds a consistent hypothesis in runtime of
$\tilde{O}((m + d) \cdot C)$.  For the same problem, our algorithm
(with the Perceptron as the weak learner) finds a consistent
hypothesis in runtime of $\tilde{O}( (m + C) \cdot 
d)$. Furthermore, if the instances are $\bar{d}$-sparse (meaning that
the number of non-zeros in each $x_i$ is at most $\bar{d}$), then the
term $d$ in our bound can be replaced by $\bar{d}$. In any case, our
bound is sometimes better and sometimes worse than the one in
\cite{clarkson2012sublinear}.  We note that we can also use
AdaBoost~\cite{freund1995desicion} on top of the Perceptron algorithm
for the same problem. It can be easily verified that the resulting
runtime will be identical to our bound. In this sense, our algorithm
can be seen as an online version of AdaBoost.

Finally, several recent works use sampling strategies for speeding up
optimization algorithms for minimizing the average loss. See for
example \cite{bengio2008adaptive,bouchard2015accelerating,zhao2014stochastic,allen2015even}.

\section{Why} \label{sec:why}

In this section we tackle the ``Why'' question, namely, why should we
prefer minimizing the maximal loss instead of the average loss. For
simplicity of presentation, throughout this section we deal with
binary classification problems with the zero-one loss, in the
realizable setting. In this context, minimizing the maximal loss to
accuracy of $\epsilon < 1$ leads to a consistent
hypothesis\footnote{Recall that a consistent hypothesis is a
  hypothesis that makes no mistakes on the training set. We also use
  the term Empirical Risk Minimization (ERM) to describe the process
  of finding a consistent hypothesis, and use $\textrm{ERM}(S)$ to
  denote any hypothesis which is consistent with a sample $S$.}. On
the other hand, minimizing the average loss to any accuracy of
$\epsilon > 1/m$ does not guarantee to return a consistent
hypothesis. Therefore, in this context, the ``why'' question becomes:
why should we find a consistent hypothesis and not be satisfied with a
hypothesis with $L_\avg(h) \le \epsilon$ for some $\epsilon > 1/m$.

In the usual PAC learning model (see \cite{MLbook} for an overview),
there is a distribution $\D$ over $\X \times \Y$ and the training
examples are assumed to be sampled i.i.d. from $\D$. The goal of the
learner is to minimize $L_{\D}(h) :=\E_{(x,y) \sim \D}[\ell(h,x,y)] =
\prob_{(x,y) \sim \D}[h(x) \neq y]$. For a fixed $h \in \cH$, the
random variable $L_\avg(h)$ is an unbiased estimator of
$L_{\D}(h)$. Furthermore, it can be shown (\citet[Section
5.1.2]{boucheron2005theory}) that with probability of at least
$1-\delta$ over the choice of the sample $S \sim \D^m$ we have that:
\begin{align*}
&\forall h \in \cH,~~ L_{\D}(h) \le L_\avg(h) + \\
&\tilde{O}\left( \sqrt{L_\avg(h)
  \frac{\textrm{VC}(\cH) - \log(\delta)}{m}} + 
 \frac{\textrm{VC}(\cH) - \log(\delta)}{m}\right) 
\end{align*}
where $\textrm{VC}(\cH)$ is the VC dimension of the class $\cH$ and
the notation $\tilde{O}$ hides constants and logarithmic terms. 

From the above bound we get that any $h$ with $L_\avg(h) = 0$
(i.e., a consistent $h$) guarantees that $L_{\D}(h) = \tilde{O}\left(
  \frac{\textrm{VC}(\cH)+ \log(1/\delta)}{m}\right)$.  However, we
will obtain the same guarantee (up to constants) if we will choose any
$h$ with $L_\avg(h) \le \epsilon$, for $\epsilon = \tilde{O}\left(
  \frac{\textrm{VC}(\cH)+ \log(1/\delta)}{m}\right)$. Based on this
observation, it can be argued that it is enough to minimize $L_\avg$
to accuracy of $\epsilon = \tilde{O}\left( \frac{\textrm{VC}(\cH)+
    \log(1/\delta)}{m}\right) > \frac{1}{m}$, because a better
accuracy on the training set will in any case get lost by the sampling
noise.

Furthermore, because of either computational reasons or high
dimensionality of the data, we often do not directly minimize the
zero-one loss, and instead minimize a convex surrogate loss, such as
the hinge-loss. In such cases, we often rely on a margin based
analysis, which means that the term $\textrm{VC}(\cH)$ is replaced by
$B^2$, where $B$ is an upper bound on the norm of the weight vector
that defines the classifier. It is often the case that the convergence
rate of SGD is of the same order, and therefore there is no added
value of solving the ERM problem over performing a single SGD pass
over the data (or few epochs over the data).  Formal arguments of this
nature were given in \cite{bousquet2008tradeoffs,shalev2008svm}.

Despite of these arguments, we show below reasons to prefer the max
loss formulation over the average loss formulation. The first reason is
straightforward: arguments that are based on worst case bounds are
problematic, since in many cases the behavior is rather different than
the worst case bounds. In subsection~\ref{sec:simpleGap} we present a
simple example in which there is a large gap between the sample
complexity of SGD and the sample complexity of ERM, and we further
show that the runtime of our algorithm will be much better than the
runtime of SGD for solving this problem. 

Next, we describe a family of problems in which the distribution from
which the training data is being sampled is a mix of ``typical''
examples and ``rare'' examples. We show that in such a case, few
``rare'' examples may be sufficient for learning a hypothesis that has
a high accuracy on both the ``typical'' and ``rare'' examples, and
therefore, it is really required to solve the ERM problem as opposed
to being satisfied with a hypothesis for which $L_\avg(h)$ is small.

\subsection{A Simple Example of a Gap} \label{sec:simpleGap} 

Consider the following distribution. Let $z_1 = (\alpha,1)$ and $z_2 =
(\alpha,-2\alpha)$ for some small $\alpha > 0$. To generate an example
$(x,y) \sim \D$, we first sample a label $y$ uniformly at random from $\{\pm
1\}$, then we set $x = y z_1$ with probability $1-\epsilon$ and set $x
= y z_2$ with probability $\epsilon$. The hypothesis class is
halfspaces: $\cH = \{ x \to \textrm{sign}(\inner{w,x}) : w \in \reals^2\}$.

The following three lemmas, whose proofs are given in the appendix,
establish the gap between the different approaches.

\begin{lemma} \label{lem:exampleERMgood}
  For every $\delta \in (0,1)$, if $m \ge \frac{2
    \log(4/\delta)}{\epsilon}$ then, with probability of at least
  $1-\delta$ over the choice of the training set, $S \sim \D^m$, any
  hypothesis in $\textrm{ERM}(S)$ has a generalization error of $0$.
\end{lemma}

\begin{lemma} \label{lem:exampleSGDbad}
  Suppose we run SGD with the hinge-loss and any $\eta > 0$ for less
  than $T = \Omega(1/(\alpha \epsilon))$ iterations. Then, with
  probability of $1-O(\epsilon)$ we have that SGD will not find a
  solution with error smaller than $\epsilon$.
\end{lemma}

\begin{lemma} \label{lem:exampleROBgood}
Running FOL (with the Perceptron as its $w$ player) takes $\tilde{O}\left(\frac{1}{\epsilon} +
\frac{1}{\alpha}\right)$ iterations.
\end{lemma}

\subsection{Typical vs. Rare Distributions} \label{sec:typicalRare}

\begin{figure}[ht]
  \centering
  \includegraphics[width=0.49\textwidth]{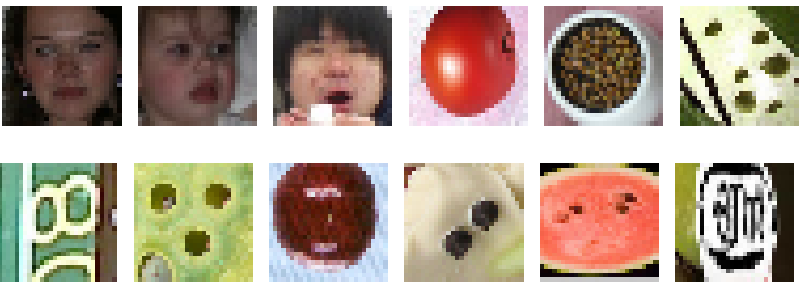}
  \caption{Top: typical positive (left) and negative (right)
    examples. Bottom: rare negative examples.} \label{fig:faces}
  \label{fig:rare}
\end{figure}

To motivate the learning setting, consider the problem of
face detection, in which the goal is to take an image crop and
determine whether it is an image of a face or not. An illustration of
typical random positive and negative examples is given in
\figref{fig:faces} (top row). By having enough training examples, we
can learn that the discrimination between face and non-face is based
on few features like ``an ellipse shape'', ``eyes'', ``nose'', and
``mouth''. 
However, from the typical examples it is hard
to tell whether an image of a watermelon is a face or not --- it has the
ellipse shape like a face, and something that looks like eyes, but it
doesn't have a nose, or a
mouth. The bottom row of \figref{fig:faces} shows some additional ``rare''
examples. 
% \begin{figure}[ht]
%   \centering
%   %\subfigure{ \includegraphics[width=0.07\textwidth]{\figurespath{}0101-img1851_80x120.png}}
%   %\subfigure{ \includegraphics[width=0.07\textwidth]{\figurespath{}0379-img7463_104x72.png}}
%   %\subfigure{ \includegraphics[width=0.07\textwidth]{\figurespath{}0116-img2121_72x84.png}}
%   \subfigure{ \includegraphics[width=0.07\textwidth]{\figurespath{}face23.png}}
%   \subfigure{ \includegraphics[width=0.07\textwidth]{\figurespath{}face26.png}}
%   \subfigure{ \includegraphics[width=0.07\textwidth]{\figurespath{}face59.png}}
%   \subfigure{ \includegraphics[width=0.07\textwidth]{\figurespath{}0625-img12572_56x40.png}}
%   \subfigure{ \includegraphics[width=0.07\textwidth]{\figurespath{}0569-img11151_28x48.png}}
%   \subfigure{ \includegraphics[width=0.07\textwidth]{\figurespath{}0364-img7163_104x140.png}}
%   \\
%   \subfigure{ \includegraphics[width=0.07\textwidth]{\figurespath{}0004-img103_84x104.png}}
%   \subfigure{ \includegraphics[width=0.07\textwidth]{\figurespath{}0007-img133_72x92.png}}
%   \subfigure{ \includegraphics[width=0.07\textwidth]{\figurespath{}0011-img143_76x144.png}}
%   \subfigure{ \includegraphics[width=0.07\textwidth]{\figurespath{}0105-img1954_36x84.png}}
%   \subfigure{ \includegraphics[angle=90,width=0.07\textwidth]{\figurespath{}0187-img3426_100x128.png}}
%   \subfigure{ \includegraphics[width=0.07\textwidth]{\figurespath{}0146-img2597_96x56.png}}
%   \caption{Top: typical positive (left) and negative (right)
%     examples. Bottom: rare negative examples.} \label{fig:faces}
%   \label{fig:rare}
% \end{figure}

Such a phenomenon can be formally described as follows. There are two
distributions over the examples, $\D_1$ and $\D_2$. Our goal is to
have an error of at most $\epsilon$ on \textbf{both} distributions,
namely, we would like to find $h$ such that $L_{\D_1}(h) \le \epsilon$
and $L_{\D_2}(h) \le \epsilon$. However, the training examples that we
observe are sampled i.i.d. from a mixed distribution, $\D = \lambda_1
\D_1 + \lambda_2 \D_2$, where $\lambda_1,\lambda_2 \in (0,1)$ and
$\lambda_1 + \lambda_2 = 1$. We assume that $\lambda_2 \ll \lambda_1$,
namely, typical examples in the training set are from $\D_1$ while
examples from $\D_2$ are rare. 

Fix some $\epsilon$. If $\lambda_2 < \epsilon$, then a hypothesis
with $L_\avg(h) \le \epsilon$ might err on most of the ``rare''
examples, and is therefore likely to have 
$L_{\D_2}(h) > \epsilon$. If we want to guarantee a good performance
on $\D_2$ we must optimize to a very high accuracy, or put another
way, we would like to minimize $L_{\max}$ instead of $L_\avg$. The
question is how many examples do we need in order to guarantee that a
consistent hypothesis on $S$ will have a small error on both $\D_1$
and $\D_2$. A naive approach is to require order of $\textrm{VC}(\cH)
/ (\lambda_2 \epsilon)$ examples, thus ensuring that we have order of 
$\textrm{VC}(\cH) / \epsilon$ examples from both $\D_1$ and
$\D_2$. However, this is a rough estimate and the real sample
complexity might be much smaller. Intuitively, we can think of
the typical examples from $\D_1$ as filtering out most of the hypotheses in $\cH$,
and the goal of the rare examples is just to fine tune the exact
hypothesis. In the example of face detection, the examples from $\D_1$
will help us figure out what is an ``ellipse like shape'', what is an
``eye'', and what is a ``mouth'' and a ``nose''. After we understand
all this, the rare examples from $\D_2$ will tell us the exact
requirement of being a face (e.g., you need an ellipse like shape and
either eyes or a mouth). We can therefore hope that the number of
required ``rare'' examples is much smaller than the number of required
``typical'' examples. This intuition is formalized in the following
theorem. 

\begin{theorem} \label{thm:typicalRare}
Fix $\epsilon,\delta \in (0,1)$, distributions $D_1,D_2$, and let $D = \lambda_1 D_1 +
  \lambda_2 D_2$ where $\lambda_1+\lambda_2 = 1, \lambda_1,\lambda_2
  \in [0,1]$, and $\lambda_2 < \lambda_1$. Define $\cH_{1,\epsilon} = \{ h \in \cH : L_{D_1}(h) \le
  \epsilon \}$ and $c = \max\{c' \in [\epsilon,1) : \forall h \in
  \cH_{1,\epsilon}, \, L_{D_2}(h) \le c' \Rightarrow L_{D_2}(h) \le
  \epsilon\}$. Then, if
\begin{align*}
m \ge \Omega&\left( \frac{\mathrm{VC}(\cH) \log(1/\epsilon) +
    \log(1/\delta)}{\epsilon} + \right. \\
&\left.\frac{\mathrm{VC}(\cH_{1,\epsilon})  \log(1/c) + \log(1/\delta)}{c\,\lambda_2}  \right)
\end{align*}
we have, with probability of at least $1-\delta$ over the sampling of
a sample $S \sim D^m$:
\[
L_{D_1}(\mathrm{ERM}(S)) \le \epsilon ~~~and~~~
L_{D_2}(\mathrm{ERM}(S)) \le \epsilon
\]
\end{theorem}

The proof of the theorem is given in the appendix. The first term in
the sample complexity is a standard VC-based sample complexity. The
second term makes two crucial improvement. First, we measure the VC
dimension of a reduced class ($\cH_{1,\epsilon}$), containing only
those hypotheses in $\cH$ that have a small error on the ``typical''
distribution. Intuitively, this will be a much smaller hypothesis
class compared to the original class. Second, we apply an analysis of
the sample complexity similar to the ``shell analysis'' of
\cite{haussler1996rigorous}, and assume that the error of all
hypotheses in $\cH_{1,\epsilon}$ on $\D_2$ is either smaller than
$\epsilon$ or larger than $c$, where we would like to think of $c$ as
being significantly larger than $\epsilon$.  Naturally, this will not
always be the case. But, \thmref{thm:typicalRare} provides data
dependent conditions, under which a much smaller number of examples
from $\D_2$ is sufficient. As a motivation, consider again
\figref{fig:rare}, and suppose $H_{1,\epsilon}$ contains conjunctions
over all subsets of the features ``has eyes'', ``has nose'', ``has
mouth'', ``has skin color''. Let $h^*$ be the conjunction of all these
4 features. It is reasonable to assume that examples in $\D_2$ lack
one of these features. Let us also assume for simplicity that each
lacking feature takes at least $1/8$ of the mass of $\D_2$. Hence, the
error of all ``wrong'' functions in $\cH_{1,\epsilon}$ on $\D_2$ is at
least $1/8$, while the error of $h^*$ is $0$. We see that in this
simple example, $c = 1/8$.

All in all, the theorem shows that a small number of ``rare'' examples
in the training set can have a dramatic effect on the performance of
the algorithm on the rare distribution $\D_2$. But, we will see this
effect only if we will indeed find a hypothesis consistent with all
(or most) examples from $\D_2$, which requires an algorithm for
minimizing $L_{\max}$ and not $L_\avg$.

\section{Robustness} \label{sec:robust}

In the previous section we have shown cases in which minimizing
$L_{\max}$ is better than minimizing $L_\avg$. However, in the
presence of outliers, minimizing $L_{\max}$ might lead to meaningless
results --- even a single outlier can change the value of $L_{\max}$
and might lead to a trivial, non-interesting, solution. 
In this section we describe two tricks for addressing this
problem. The first trick replaces the original sample with a new
sample whose examples are sampled from the original sample. The second
trick relies on slack variables. We note that these tricks are not
new and appears in the literature in various forms. See for example
\cite{huber2009robust,maronnarobust}. The goal of this section is
merely to show how to apply known tricks to the max loss problem.

Recall that in the previous section we have shown that a small amount
of ``rare'' examples can have a dramatic effect on the performance of
the algorithm on the ``rare'' distribution. Naturally, if the number
of outliers is larger than the number of rare examples we cannot hope
to enjoy the benefit of rare examples. Therefore, throughout this
section we assume that the number of outliers, denoted $k$, is smaller
than the number of ``rare'' examples, which we denote by $m_2$.

\subsection{Sub-sampling with repetitions}

The first trick we consider is to simply take a new sample of $n$
examples, where each example in the new sample is sampled independently
according to the uniform distribution over the original $m$
examples. Then, we run our algorithm on the obtained sample of $n$
examples. 

Intuitively, if there are $k$ outliers, and the size of the new sample
is significantly smaller than $m/k$, then there is a good chance that
no outliers will fall into the new sample. On the other hand, we want
that enough ``rare'' examples will fall into the new sample. The
following theorem, whose proof is in the appendix, shows for which
values of $k$ and $m_2$ this is possible.
\begin{theorem} \label{thm:subsampleRobust}
Let $k$ be the number of outliers, $m_2$ be the number of rare
examples, $m$ be the size of the original sample, and $n$ be the size
of the new sample. Assume that $m \ge 10k$.  Then, the probability that the new sample contains
outliers and/or does not contain at least $m_2/2$ rare examples is at
most $0.01 + 0.99 k n / m + e^{-0.1\,n m_2 /m}$.
\end{theorem}

For example, if $n = m / (100 k)$ and $m_2 \ge 1000\,\log(100)\,k$,
then the probability of the bad event is at most $0.03$. 

% Another option is to take several samples, for each one define a
% maximal loss, and minimize the average over them. On samples for which
% we have outliers, we'll get an optimal loss of $1$, and on ``good''
% samples we'll get an optimal loss of $0$. If we have probability of
% $\alpha$ for ``good'' sample, we should be able to find a solution
% with error close to $\alpha$. Suppose that $\alpha = 0.25$ and we get
% a solution with error of $0.5$. Then, on half of the samples we have
% zero error. If at least one of them contain enough ``rare'' examples
% we'll have a good solution. 
% \todo{Formalize and show how to adapt the algorithm to work on such a case.}

\subsection{Slack variables}
Another common trick, often used in the SVM literature, is
to introduce a vector of slack variables, $\xi \in \reals^m$, such that
$\xi_i > 0$ indicates that example $i$ is an outlier. We first
describe the ideal version of outlier removal. Suppose we restrict
$\xi_i$ to take values in $\{0,1\}$, and we restrict the number of
outliers to be at most $K$. Then, we can write the following
optimization problem:
\begin{align*}
\min_{w \in \cW, \xi \in \reals^m} ~\max_{i \in [m]} ~ &
(1-\xi_i) \, \ell(w,x_i,y_i)  ~~\textrm{s.t.} \\
&\xi \in \{0,1\}^m,~
\|\xi\|_1 \le K ~.
\end{align*}
This optimization problem minimizes the max loss over a subset of
examples of size at least $m-K$. That is, we allow the algorithm to
refer to at most $K$ examples as outliers. 

Note that the above problem can be written as a max-loss minimization:
\begin{align*}
&\min_{\bar{w} \in \bar{\cW}} \max_i \bar{\ell}(\bar{w},x_i,y_i)
~~~\textrm{where} \\
& \bar{\cW} = \{(w,\xi) : w \in \cW, \xi \in \{0,1\}^m, \|\xi\|_1 \le
K\} ~~\textrm{and}~~ \\
&\bar{\ell}((w,\xi),x_i,y_i) = (1-\xi_i)\ell(w,x_i,y_i)
\end{align*}
We can now apply our framework on this modified problem. The $p$
player remains as before, but now the $\bar{w}$ player has a more
difficult task. To make the task easier we can perform several
relaxations. First, we can replace the non-convex constraint $\xi \in
\{0,1\}^m$ with the convex constraint $\xi \in [0,1]^m$. 
Second, we can replace the multiplicative slack with an additive
slack, and re-define: $\bar{\ell}((w,\xi),x_i,y_i) = \ell(w,x_i,y_i) -
\xi_i$. This adds a convex term to the loss function, and therefore,
 if the original loss was convex we end up with a convex loss. The new
 problem can often be solved by combining gradient updates with
 projections of $\xi$ onto the set $\xi \in [0,1]^m, \|\xi\|_1 \le
 K$. For efficient implementations of this projection see for example
 \cite{duchi2008efficient}. We can further replace the constraint $\|\xi\|_1 \le K$ with a
 constraint of $\|\xi\|_2^2 \le K$, because projection onto the
 Euclidean ball is a simple scaling, and the operation can be done
 efficiently with an adequate data structure (as described, for
 example, in \cite{shalev2011pegasos}).

\section{Experiments} \label{sec:experiments}
\input{experiments.tex}

\paragraph{Acknowledgements:} S. Shalev-Shwartz is supported by 
ICRI-CI and by the European Research Council (TheoryDL project).

%%% Local Variables:
%%% mode: latex
%%% TeX-master: "RobICML"
%%% End:

%% file: experiments.tex
\newcommand{\plh}{{\mkern-1mu\times\mkern-1mu}}

In this section we demonstrate several merits of our approach on the
well studied problem of face detection.  Detection problems in general
have a biased distribution as they are often expected to detect few
needles in a haystack.  Furthermore, a mix of typical and rare
distributions is to be expected.  For example, users of smartphones
won't be in the same continent as the manufacturers who collect data
for training.  This domain requires weighting of examples, and
therefore is a good playground to examine our algorithm.

To create a dataset we downloaded $30k$ photos from Google images that
are tagged with ``face''.  We then applied an off-the-shelf face
detector, and it found $32k$ rectangles that aligned on faces.  This
was the base of our positive examples.  For negative examples we
randomly sampled $250k$ rectangles in the same images that do not
overlap faces.  Each rectangle was cropped and scaled to $28 \plh 28$
pixels.  Using a fixed size simplifies the experiments so we can focus
on the merits of our method rather than justify various choices that
are not relevant here, such as localization and range of scale. 

Recall that our FOL algorithm relies on an online algorithm as the $w$
player. In the experiments, $w$ is the vector containing all the
weights of a convolutional neural network (CNN) with a variant of the well
known LeNet architecture. The layers of the network are as
follows. Convolution with a $5\plh 5$ kernel, stride of $1$, and $40$
output channels, followed by ReLU and max-pooling.  Then a convolution
with a $5 \plh 5$ kernel, stride of $1$, and $80$ output channels,
followed by ReLU and max-pooling.  Then a convolution with a $7\plh 7$
kernel, stride of $1$, and $160$ output channels, followed by
ReLU. Finally, a linear prediction over the resulting $160$ channels
yields the binary prediction for the input $28 \plh 28$ image crop.
Overall the model has $710,642$ weights. We denote by $\cH$ the resulting hypothesis class. 

In the comparison, we focus on the case in which the data is
realizable by $\cH$. To guarantee that, we first used vanilla SGD to
find a network in $\cH$.  We then kept only samples that were labeled
correctly by the network. This yielded $28k$ positive examples and
$246k$ negative examples.  This set was then randomly mixed and split
$9:1$ for train and test sets.

For the $w$ player we used the SGD algorithm with Nesterov's momentum,
as this is a standard solver for learning convolutional neural
networks. The parameters we used are a batch size of $64$, an $\ell_2$
regularization of $0.0005$, momentum parameter of $0.9$, and a
learning rate of $\eta_t = 0.01 (1+0.0001 \,t)^{-0.75}$.  We used the
logistic loss as a surrogate loss for the classification error.

We performed two experiments to highlight different properties of
our algorithm.  The first experiment shows that FOL is much faster than SGD, and this is reflected both in train and test errors.  The second experiment compares FOL to the application of AdaBoost on top of the same base learner.

\input{combined_figures.tex}

\paragraph{Experiment 1: Convergence speed}
% SGD vs. boost SGD. The current graph is great. Explain that
% $1/0.0005^2$ is large. Explain also the intuition that the
% probability to see relevant example is $140/280000$.
In this experiment we show that FOL is faster than SGD. \figref{fig:speed} shows
the train and test errors of SGD and FOL. Both models were initialized
with the same randomly selected weights.  As mentioned before, FOL
relies on SGD as its $w$ player.  Observe that FOL essentially solves
the problem (zero training error) after $30$ epochs, whereas SGD did
not converge to a zero training error even after $14,000$ epochs and
achieved $0.1313\%$ error.  While the logarithmic scale in the figure
shows that SGD is still improving, it is doing so at a decreasing
rate.  This is reflected by our theoretical analysis.  
To understand why SGD slows down, observe that when the error of SGD
is as small as here ($0.13\%$), only one example in $769$ is
informative.  Even when at classification error of $0.4\%$ (left-side
of the graph), only $4$ in $1,000$ examples are informative.  Hence,
even with batch size of $64$, SGD picks one useful sample only once
every fifteen iterations.  FOL expects an average of $32$ useful
samples in every iteration so every iteration is informative.  In our
case, since the training set size is $246k$, only $984$ examples are
informative and FOL makes sure to focus on these rather than waste
time on solved examples.

As can be seen in \figref{fig:speed}, the faster convergence of FOL on
the training set is also translated to a better test error. Indeed,
FOL achieves a test error of $0.14\%$ (after $27$ epochs) whereas even
after $14k$ epochs SGD results $0.35\%$ error.

\paragraph{Experiment 2: Comparison to AdaBoost}
As mentioned in \secref{sec:related}, FOL can be seen as an online
version of AdaBoost. Specifically, we can apply the AdaBoost algorithm
while using SGD as its weak learner. For concreteness and
reproducibility of our experiment, we briefly describe the resulting
algorithm. We initialize a uniform distribution over the $m$ examples,
$p = (1/m,\ldots,1/m)$. At iteration $t$ of AdaBoost, we run one epoch
of SGD over the data, while sampling examples according to $p$. Let
$h_t$ be the resulting classifier. We then calculate $h_t(x_i)$ over
all the $m$ examples, calculate the averaged zero-one error of $h_t$,
with weights based on $p$, define a weight $\alpha_t = 0.5
\log(1/\epsilon_t - 1)$, and update $p$ such that $p_i \propto p_i
\exp(-\alpha_ty_i h_t(x_i))$. We repeat this process for $T$
iterations, and output the hypothesis $h(x) = \sign( \sum_{t=1}^T
\alpha_t h_t(x))$.

Observe that each such iteration of AdaBoost is equivalent to $2$
epochs of our algorithm. In \figref{fig:boost} we show the train error of
AdaBoost and FOL as a function of the number of epochs over the data.
The behavior on the test set shows a similar trend.  As can be seen in
the figure, AdaBoost finds a consistent hypothesis after $20$ epochs,
while FOL requires $27$ epochs to converge to a consistent
hypothesis. However, once FOL converged, its last hypothesis has a
zero training error. In contrast, the output hypothesis of AdaBoost is
a weighted majority of $T$ hypotheses ($T=10$ in this case). It follows
that at prediction time, applying AdaBoost's predictor is $10$ times
slower than applying FOL's predictor. Often, we prefer to spend more
time during training, for the sake of finding a hypothesis which can
be evaluated faster at test time. While based on our theory, the
output hypothesis of FOL should also be a majority of $\log(m)$
hypotheses, we found out that in practice, the last hypothesis of FOL
converges to a zero classification error at almost the same rate as
the majority classifier.

%%% Local Variables:
%%% mode: latex
%%% TeX-master: "RobICML"
%%% End:

%% file: combined_figures.tex
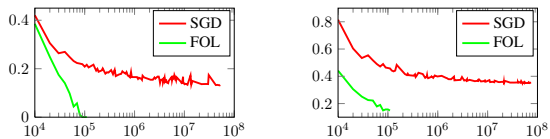
\begin{figure}
\begin{center}
  \begin{minipage}[b]{0.23\textwidth}
    \input{FOL-vs-SGD-train-B5.tex}
  \end{minipage}
  \begin{minipage}[b]{0.23\textwidth}
    \input{FOL-vs-SGD-test-B5.tex}
  \end{minipage}

\end{center}
\caption{Comparing the error percentage of FOL vs. SGD as a function of the number of iterations. 
Left: Train Error. Right: Test Error.} 
\label{fig:speed}
\end{figure}

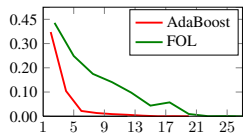
\begin{figure}
  \begin{center}
      \input{FOL_vs_AdaBoost.tex}
  \end{center}
\caption{Train error of FOL vs. AdaBoost as a function of the number of epochs.} 
\label{fig:boost}
\end{figure}

\iffalse
\begin{figure*}[ht!]
  \centering
%
  %\input{figures/FOL-vs-SGD_D1train.tex}
  \begin{minipage}[b]{0.24\textwidth}
    \input{figures/final/FOL-vs-SGD-train-B5.tex}
  \end{minipage}
  \quad
%
  %\input{figures/FOL-vs-SGD_D1test.tex}
  \begin{minipage}[b]{0.24\textwidth}
    \input{figures/final/FOL-vs-SGD-test-B5.tex}
  \end{minipage}
%
  %\begin{minipage}[b]{0.24\textwidth}
  %  %\input{figures/FOL_SGD_generalization.tex}
  %  \input{figures/final/FOL_SGD_generalization.tex}
  %\end{minipage}
%
  \begin{minipage}[b]{0.24\textwidth}
    \input{figures/final/FOL_vs_AdaBoost.tex}
  \end{minipage}
\end{figure*}
\fi

%%% Local Variables:
%%% mode: latex
%%% TeX-master: "RobICML"
%%% End:

%% file: FOL-vs-SGD-train-B5.tex
% This file was created by matplotlib2tikz v0.5.4.
% The lastest updates can be retrieved from
% 
% https://github.com/nschloe/matplotlib2tikz
% 
% where you can also submit bug reports and leavecomments.
% 
\begin{tikzpicture}[scale=0.6]

\begin{axis}[
xmin=10000, xmax=100000000,
ymin=0, ymax=0.45,
xmode=log,
axis on top,
width=6cm,
height=4cm,
legend cell align={left},
legend entries={{SGD},{FOL}}
]
\addplot [very thick, red]
table {%
10000 0.420731
20000 0.305691
30000 0.263821
40000 0.269512
50000 0.247968
60000 0.231708
70000 0.223577
80000 0.221138
90000 0.218699
100000 0.208537
110000 0.21626
120000 0.205285
150000 0.216667
160000 0.203658
170000 0.192276
180000 0.199594
190000 0.187398
220000 0.193089
230000 0.20122
260000 0.179675
280000 0.185366
290000 0.184553
300000 0.184146
340000 0.194309
370000 0.177236
400000 0.195528
420000 0.169512
490000 0.176423
540000 0.178862
570000 0.189025
580000 0.187398
660000 0.179675
690000 0.186179
730000 0.171951
750000 0.171138
810000 0.164634
850000 0.186992
880000 0.165447
1120000 0.165447
1170000 0.158536
1190000 0.161382
1220000 0.160569
1230000 0.162602
1320000 0.160163
1380000 0.156911
1410000 0.154065
1450000 0.171138
1550000 0.164228
2100000 0.154472
2330000 0.154472
2510000 0.163821
2610000 0.158943
2890000 0.16748
3060000 0.16748
3170000 0.155285
3250000 0.141057
3350000 0.151626
3690000 0.157724
3720000 0.142683
4000000 0.164228
4070000 0.146342
4190000 0.150406
4320000 0.152033
4380000 0.143902
4730000 0.140244
4840000 0.149187
5450000 0.163821
5630000 0.155285
5650000 0.151626
5750000 0.158943
5810000 0.165041
6150000 0.158943
6300000 0.173577
6730000 0.15691
6840000 0.151626
7030000 0.164228
8300000 0.148374
8410000 0.147155
8810000 0.140244
8970000 0.14065
9420000 0.166667
9470000 0.141464
10000000 0.136992
10070000 0.138211
10610000 0.137398
11040000 0.136585
11120000 0.135772
11530000 0.13374
13280000 0.172358
13750000 0.134553
14400000 0.149594
22250000 0.136179
23020000 0.136585
27220000 0.136992
29890000 0.136585
30760000 0.173984
42670000 0.13252
45920000 0.133333
51220000 0.131301
52550000 0.13252
};
\addplot [very thick, green]
table {%
10000 0.385366
20000 0.248781
30000 0.174797
40000 0.14065
50000 0.097561
60000 0.0439024
70000 0.0573171
80000 0.00934959
90000 0.000406504
100000 0.000813009
110000 0
};

\end{axis}

\end{tikzpicture}

%% file: FOL-vs-SGD-test-B5.tex
% This file was created by matplotlib2tikz v0.5.4.
% The lastest updates can be retrieved from
% 
% https://github.com/nschloe/matplotlib2tikz
% 
% where you can also submit bug reports and leavecomments.
% 
\begin{tikzpicture}[scale=0.6]

\begin{axis}[
xmin=10000, xmax=100000000,
ymin=0.1, ymax=0.9,
xmode=log,
axis on top,
width=6cm,
height=4cm,
legend cell align={left},
legend entries={{SGD},{FOL}}
]
%\addplot [very thick, green!50.0!black]
\addplot [very thick, red]
table {%
10000 0.814857
20000 0.60385
30000 0.533953
40000 0.553979
50000 0.517143
60000 0.489791
70000 0.471909
80000 0.489966
90000 0.464252
110000 0.457639
120000 0.441816
130000 0.436287
140000 0.441603
150000 0.483731
220000 0.416125
240000 0.408844
250000 0.419169
260000 0.408285
330000 0.404235
350000 0.414545
360000 0.394546
370000 0.399869
410000 0.425844
500000 0.41655
560000 0.388486
590000 0.393841
600000 0.39751
610000 0.410212
620000 0.394225
640000 0.38587
680000 0.389797
740000 0.395872
750000 0.399055
810000 0.388235
870000 0.405933
980000 0.398635
1200000 0.403315
1470000 0.388182
1510000 0.387243
1640000 0.379005
1850000 0.378323
2120000 0.397571
2160000 0.37035
2350000 0.37133
2860000 0.373657
3060000 0.394717
3350000 0.371892
3380000 0.369257
3710000 0.370566
3860000 0.378912
3870000 0.36628
4320000 0.374124
5550000 0.364053
6040000 0.366126
6430000 0.365044
6450000 0.374502
6690000 0.377687
7340000 0.362238
8330000 0.364415
8800000 0.369333
8880000 0.360748
9150000 0.360138
10880000 0.371973
11210000 0.357272
12540000 0.359549
13640000 0.356223
13820000 0.368117
14510000 0.365532
16450000 0.357698
17550000 0.357336
18710000 0.358898
20870000 0.360185
21180000 0.355105
21260000 0.359495
21330000 0.355154
21350000 0.353452
22610000 0.365424
28340000 0.353862
31500000 0.354014
32890000 0.35381
34260000 0.353924
40520000 0.354786
40550000 0.35855
40940000 0.355381
42940000 0.355229
44270000 0.356686
45670000 0.353281
47020000 0.364923
49250000 0.35173
49280000 0.353011
51220000 0.350297
54020000 0.358335
55330000 0.350872
56230000 0.354981
58580000 0.350628
65110000 0.350697
66700000 0.356362
67810000 0.353936
70340000 0.350454
74060000 0.355983
};
\addplot [very thick, green]
table {%
10000 0.441463
20000 0.305691
30000 0.250813
40000 0.22561
50000 0.222358
60000 0.178862
70000 0.189837
80000 0.149593
90000 0.156911
100000 0.156911
110000 0.149187
%120000 0.153252
%130000 0.138211
%140000 0.138211
%150000 0.149187
%160000 0.145528
%170000 0.152846
%180000 0.130894
%190000 0.14187
%200000 0.138211
%210000 0.149187
%220000 0.152845
%230000 0.14187
%240000 0.138211
%250000 0.149187
%260000 0.134553
%270000 0.127236
%280000 0.152845
%290000 0.138211
%300000 0.130894
%310000 0.149187
%320000 0.145528
%330000 0.14187
%340000 0.138211
%350000 0.156504
%360000 0.14187
%370000 0.149187
%380000 0.152846
%390000 0.160163
%400000 0.145528
};
\end{axis}

\end{tikzpicture}

%% file: FOL_vs_AdaBoost.tex
% This file was created by matplotlib2tikz v0.5.4.
% The lastest updates can be retrieved from
% 
% https://github.com/nschloe/matplotlib2tikz
% 
% where you can also submit bug reports and leavecomments.
% 
\begin{tikzpicture}[scale=0.6]

\begin{axis}[
xmin=1, xmax=27,
ymin=0, ymax=0.45,
axis on top,
width=6cm,
height=4cm,
ytick={0,0.1,0.2,0.3,0.4,0.45},
yticklabels={0.00,0.10,0.20,0.30,0.45},
xtick={1,3,5,7,9,11,15,19,23,27},
xtick={1,5,9,13,17,21,25},
legend entries={{AdaBoost},{FOL}},
legend cell align={left}
]
\addplot [very thick, red]
table {%
  2   0.347515
  4   0.104092
  6   0.021872
  8   0.013366
  10  0.008911
  12  0.006075
  14  0.002430
  16   0.000000
  18  0.000405
  20  0.000000
};
\addplot [very thick, green!50.0!black]
table {%
2.5 0.385366
5 0.248781
7.5 0.174797
10 0.14065
12.5 0.097561
15 0.0439024
17.5 0.0573171
20 0.00934959
22.5 0.000406504
25 0.000813009
27.5 0
};
\path [draw=black, fill opacity=0] (axis cs:0,1)
--(axis cs:12,1);

\path [draw=black, fill opacity=0] (axis cs:1,0)
--(axis cs:1,0.45);

\path [draw=black, fill opacity=0] (axis cs:0,0)
--(axis cs:12,0);

\path [draw=black, fill opacity=0] (axis cs:0,0)
--(axis cs:0,0.45);

\end{axis}

\end{tikzpicture}

%% file: RobProofs.tex
\section{Proof of \thmref{thm:mainAlgo}} \label{sec:proofMain}

\subsection{Background}

\paragraph{Bernstein's type inequality for martingales:}

A sequence $B_1,\ldots,B_T$ of random variables is Markovian if for
every $t$, given $B_{t-1}$ we have that $B_t$ is independent of
$B_1,\ldots,B_{t-2}$.  A sequence $A_1,\ldots,A_T$ of random variables
is a martingale difference sequence with respect to $B_1,\ldots,B_T$
if for every $t$ we have $\E[A_t|B_1,\ldots,B_t]=0$. 

\begin{lemma}[{\citet[Lemma C.3]{hazan2011beating} and \citet[Theorem
  2.1]{fan2012hoeffding}}] \label{lem:bernstein}
Let $B_1,\ldots,B_T$ be a Markovian sequence and let $A_1,\ldots,A_T$
be a martingale difference sequence w.r.t. $B_1,\ldots,B_T$. Assume
that for every $t$ we have $|A_t| \le V$ and $\E[A_t^2 |
B_1,\ldots,B_t] \le s$. Then, for every
$\alpha > 0$ we have
\[
\prob\left(\frac{1}{T} \sum_{t=1}^T A_t \ge \alpha \right) \le
\exp\left( - T \frac{\alpha^2/2}{s + \alpha V/3}\right)
\]
In particular, for every $\delta \in (0,1)$, if 
\[
T \ge \frac{2(s+\alpha V/3) \log(1/\delta)}{\alpha^2} ~,
\]
then with probability of at least $1-\delta$ we have that $\frac{1}{T}
\sum_{t=1}^T A_t \le \alpha$. 
\end{lemma}

\paragraph{The EG algorithm:} Consider a sequence of vectors,
$z_1,\ldots,z_T$, where every $z_t \in \reals^m$. Consider the
following sequence of vectors, parameterized by $\eta > 0$.
The first vector is $\tilde{q}_1 = (1,\ldots,1) \in \reals^m$
and for $t \ge 1$ we define $\tilde{q}_{t+1}$ to be such that:
\[
\forall i \in [m],~~ \tilde{q}_{t+1,i}  = \tilde{q}_{t,i} \exp(-\eta
  z_{t,i}) ~.
\]
In addition, for every $t$ define $q_t = \tilde{q} / (\sum_{i=1}^m
\tilde{q}_i) \in \simp_m$. The algorithm that generates the above
sequence is known as the EG algorithm \cite{KivinenWa97}. 
\begin{lemma}[Theorem 2.22 in \cite{shalev2011online}] \label{lem:EG}
Assume that $\eta z_{t,i} \ge -1$ for every $t$ and $i$. Then, for
every $u \in \simp_m$ we have:
\[
\sum_{t=1}^T \inner{q_t - u , z_t} ~\le~ \frac{\log(m)}{\eta} + \eta
\sum_{t=1}^T \sum_{i=1}^m q_{t,i} z_{t,i}^2 ~.
\]
\end{lemma}

\subsection{Proof}

To simplify our notation we denote $\ell_i(w_t) =
\ell(w_t,x_i,y_i)$. We sometimes omit the time index $t$ when it is
clear from the context (e.g., we sometime use $q_i$ instead of
$q_{t,i}$). 

\subsubsection{The $w$ player}
By our  assumption that $C/T \le
\epsilon/8$ we have that, for every $i_1,\ldots,i_T$,
\begin{equation} \label{eqn:wRegret}
\frac{1}{T} \sum_{t=1}^T \ell_{i_t}(w_t) ~\le~ \epsilon/8
\end{equation}

\subsubsection{The $p$ player}

Recall that $p_i = \frac{1}{2m} + \frac{q_i}{2}$. Note that, for every $i$, 
\[
\frac{1}{p_i} \le 2m ~~~\textrm{and}~~~ \frac{q_i}{p_i} \le 2
\]

Define $z_t =
- \frac{\ell_{i_t}(w_t)}{p_{i_t}} e_{i_t}$.  Observe that the $p$ player applies
the EG algorithm w.r.t. the sequence
$z_1,\ldots,z_T$. Since $z_{t,i} \ge -2m$ we obtain from
\lemref{lem:EG} that if $\eta \le 1/(2m)$ then, for every $u \in \simp_m$,
\begin{align*}
&\frac{1}{T} \sum_{t=1}^T \inner{q_t -u ,z_t} ~\le~ 
\frac{\log(m)}{\eta T} 
+ \frac{\eta}{T} \sum_{t=1}^T \sum_{i=1}^m q_{t,i} z_{t,i}^2 \\
&\le \frac{\log(m)}{\eta T} 
+ \frac{\eta}{T} \sum_{t=1}^T q_{t,i_t}
\frac{\ell_{i_t}(w_t)^2}{p_{t,i_t}^2}  \\
&\le \frac{\log(m)}{\eta T} 
+ \frac{\eta }{T} \sum_{t=1}^T 4m \ell_{i_t}(w_t)^2 \\
&\le \frac{\log(m)}{\eta T} 
+ \frac{\eta 4 m}{T} \sum_{t=1}^T \ell_{i_t}(w_t) \\
&\le \epsilon/8 
+ \frac{2}{T} \sum_{t=1}^T \ell_{i_t}(w_t) ~,
\end{align*}
where in the last inequality we used $\eta = 1/(2m)$ and $T = \Omega(m \log(m) / \epsilon)$.
Rearranging, and combining with \eqref{eqn:wRegret} we obtain
\begin{equation} \label{eqn:pRegret}
\frac{1}{T} \sum_{t=1}^T \inner{u, \frac{1}{p_{t,i_t}}
  \ell_{i_t}(w_t) e_{i_t}} ~\le~ 
\frac{1}{T} \sum_{t=1}^T \left(\frac{q_{t,i_t}}{p_{t,i_t}} + 2\right) \ell_{i_t}(w_t) +
\epsilon/8 ~\le~ 
\frac{4}{T} \sum_{t=1}^T \ell_{i_t}(w_t) +
\epsilon/8 \le \frac{5 \epsilon}{8} ~.
\end{equation}

\subsubsection{Measure concentration}
Note that, if $u=e_i$, then
\[
\E[  \inner{u, \frac{1}{p_{t,i_t}}
  \ell_{i_t}(w_t) e_{i_t}}^2 | q_t,w_t] = \sum_{j=1}^m \frac{p_{t,j}}{p_{t,j}^2}
\ell_j(w_t)^2 u_j \le \frac{1}{p_{t,i}} \le 2 m  ~.
\]
Define the martingale difference sequence $A_1,\ldots,A_T$ where $A_t
= \ell_i(w_t) - \inner{u, \frac{1}{p_{t,i_t}} \ell_{i_t}(w_t)
  e_{i_t}}$. We have that $|A_t| \le (2m+1)$ and $\E[ A_t^2 | q_t,w_t]
\le 2m$. Therefore, the conditions of \lemref{lem:bernstein} holds and
we obtain that if $T \ge 6m \log(m/\delta) /
(\epsilon/8)^2$ then with probability of at least $1-\delta/m$ we have
that $\frac{1}{T} \sum_t A_t \le
\epsilon/8$. Applying a union
bound over $i \in [m]$ we obtain that with probability of at least
$1-\delta$ it holds that
\[
\forall i \in [m],~~ \frac{1}{T} \sum_t \ell_i(w_t) ~\le~ \frac{1}{T} \sum_t \inner{e_i, \frac{1}{p_{i_t}} \ell_{i_t}(w_t)
  e_{i_t}} + \epsilon/8 ~.
\] 
Combining with \eqref{eqn:pRegret} we obtain that, with probability of
at least $1-\delta$, 
\[
\forall i \in [m],~~ \frac{1}{T} \sum_{t=1}^T \ell_i(w_t) ~\le~ \frac{6
  \epsilon}{8} ~.
\]
Finally, relying on Bernstein's inequality (see Lemma B.10 in
\cite{MLbook}), it is not hard to see that if $k =
\Omega(\log(m/\delta)/\epsilon)$ then, with probability of at least
$1-\delta$ we have that
\[
\forall i \in [m],~~ \frac{1}{k} \sum_{j=1}^k \ell_i(w_{t_j}) \le 
\frac{1}{T} \sum_{t=1}^T \ell_i(w_t) + \frac{\epsilon}{4} ~,
\]
and this concludes our proof.

\section{Proofs of Lemmas in \secref{sec:simpleGap}}

\begin{proof}[\textbf{Proof of \lemref{lem:exampleERMgood}}]
There are only $4$ possible examples, so an ERM will have a
generalization error of $0$ provided we see all the $4$ examples. 
By a simple direct calculation together with the union bound over the
$4$ examples it is easy to verify that the
probability not to see all the examples is at most
$4 (1-\epsilon/2)^m \le 4 e^{-m \epsilon/2}$, and the claim follows. 
\end{proof}

\begin{proof}[\textbf{Proof of \lemref{lem:exampleSGDbad}}]
For SGD, we can assume (due to symmetry) that $y$ is always
$1$. Therefore, there are only two possible examples $z_1,z_2$. 
With probability $1-\epsilon$ the first examples is $z_1$. Also, 
$w_T$ has always the form
\[
\eta (k z_1 + r z_2) = \eta ( (k+r) \alpha, k - 2 r \alpha ) ~,
\]
where $k$ is the number of times we had a margin error on $z_1$ and
$r$ is the number of times we had a margin error on $z_2$. 
To make sure that $\inner{w_T,z_2} > 0$ we must have that
\begin{align}
&(k+r)\alpha^2 - 2(k - 2 r \alpha)\alpha > 0  ~~\Rightarrow~~ 
r > k \, \frac {2 \alpha - \alpha^2}{5\alpha^2} \approx k \frac{2}{5 \alpha}
\end{align}
Note that the first example is $z_1$ with probability of $1-\epsilon$,
hence we have that $k \ge 1$ with probability of at least
$1-\epsilon$. In addition, $r$ is upper bounded by the number of times
we saw $z_2$ as the example, and by Chernoff's bound we have that that
the probability that this number is greater than $2 m \epsilon$ is at
most $e^{-\epsilon/3} \approx (1-\epsilon/3)$.  Therefore, with
probability of $1-O(\epsilon)$ we have the requirement that $m$ must
be at least $\Omega(1/(\alpha \epsilon))$, which concludes our proof.
\end{proof}

\begin{proof}[\textbf{Proof of \lemref{lem:exampleROBgood}}]
We have shown that $m = 1/\epsilon$ examples suffices. Specifying our general
analysis to classification with the zero-one loss, it suffices to ensure that the
regret of both players will be smaller than $1/2$. The regret of the
sampling player is bounded by $O(m \log(m))$. As for the halfspace
player, to simplify the derivation, lets use the Perceptron as the
underlying player. It is easy to verify that the vector $w_T$
has the form $k z_1 + r z_2 = ((k+r) \alpha, k - 2 r
\alpha ) $, for some integers $k,r$. Lets consider two regimes. The
first is the first time when $r,k$ satisfies $\inner{w_T,z_2} > 0$. As
we have shown before, this happens when $r$ is roughly $2k/(5
\alpha)$.  Once this happens we also have that
\[
\inner{w_T,z_1} = ((k+r)\alpha^2 + k - 2r \alpha) \approx (k -
2r \alpha) \stackrel{\approx}{>}  4 k /5 > 0 ~,
\]
So, the Perceptron will stop making changes and will give us an
optimal halfspace. Next, suppose that we have a pair $r,k$ for which
$\inner{w_T,z_2} \le 0$. If we now encounter $z_2$ then we increase
$r$. If we encounter $z_1$ then
\[
\inner{w_T,z_1} \approx (k -
2r \alpha)  \stackrel{\approx}{>}  4 k /5 > 0 ~,
\]
so we'll not increase $k$. Therefore, $k$ will increase only up to a
constant, while $r$ will continue to increase until roughly $2k/(5
\alpha)$, and then the Perceptron will stop making updates. This
implies that the mistake bound of the Perceptron is bounded by $O(1/\alpha)$,
which concludes our proof.
\end{proof}

\section{Proof of \thmref{thm:typicalRare}}

We can think of the ERM algorithm as following the following three
steps. First, we sample $(i_1,\ldots,i_m) \in \{1,2\}^m$, where $\prob[i_r = j] =
\lambda_j$. Let $m_1$ be the number of indices for which $i_r = 1$ and
let $m_2 = m - m_1$. Second, we sample $S_1 \sim D_1^{m_1}$, and
define $\hat{\cH}_1$ to be all hypotheses in $\cH$ which
  are consistent with $S_1$. Last, we sample $S_2 \sim D_2^{m_2}$ and
  set the output hypotheses to be some hypothesis in
  $\hat{\cH}_1$ which is consistent with $S_2$. 

The proof relies on the following three claims, where we use $C$ to
denote a universal constant:
\begin{itemize}
\item \textbf{Claim 1:} With probability of at least $1-\delta/3$ over
  the choice of $(i_1,\ldots,i_m)$ we
  have that both $m_1 \ge \lambda_1 m / 2$  and $m_2 \ge
\lambda_2 m / 2$. 
\item \textbf{Claim 2:} Assuming that $m_1 \ge C\,\left(
    \frac{\mathrm{VC}(\cH) \log(1/\epsilon) + \log(1/\delta)}{\epsilon}  \right) $, then with
  probability of at least $1-\delta/3$ over the choice of $S_1$ we have that $\hat{\cH}_1 
  \subseteq \cH_{1,\epsilon}$. 
\item \textbf{Claim 3:} Assume that $m_2 \ge C\,\left(
    \frac{\mathrm{VC}(\cH_{1,\epsilon})\log(1/c) + \log(1/\delta)}{c}  \right)$, then with
  probability of at least $1-\delta/3$ over the choice of $S_2$, any  hypothesis  in
  $\cH_{1,\epsilon}$  which is consistent with $S_2$ must have
  $L_{D_2}(h) \le c$. 
\end{itemize}
Claim 1 follows directly from  Chernoff's bound, while Claim 2-3
follows directly from standard VC bounds (see for example
\citet[Theorem 6.8]{MLbook}).  

Equipped with the above three claims we are ready to prove the
theorem. First, we apply the union bound to get that with probability of at least
$1-\delta$, the statements in all the above three claims hold. This
means that $\hat{\cH}_1 \subseteq \cH_{1,\epsilon}$ hence
$L_{D_1}(\textrm{ERM}(S)) \le \epsilon$. It also means that
$\textrm{ERM}(S)$ must be in $\cH_{1,\epsilon}$, and therefore from
the third claim and the assumption in the theorem we have that
$L_{D_2}(\textrm{ERM}(S)) \le \epsilon$ as well, which concludes our
proof.

\section{Proof of \thmref{thm:subsampleRobust}}
The probability that all of the outliers do not fall into the sample
of $n$ examples is
\[
(1-k/m)^n \ge 0.99\, e^{-k n / m} ~.
\]
Therefore, the probability that at least one outlier falls into the
sample is at most 
\[
1 - 0.99\,e^{-k n /m} \le 1 - 0.99 (1 - k n / m) = 0.01 + 0.99 k n / m
\]
On the other hand, the expected number of rare examples in the sample
is $n m_2/m $ and by Chernoff's bound, the probability that less than 
half of the rare examples fall into the sample is at most $\exp(-
0.1\,n m_2 /m)$. Applying the union bound we conclude our proof.

%%% Local Variables:
%%% mode: latex
%%% TeX-master: "RobICML"
%%% End: